\documentclass{article}
\usepackage{ijcai25}

\usepackage{times}
\usepackage[small]{caption}
\usepackage{amsthm}
\usepackage{algorithm}
\usepackage{algorithmic}
\usepackage[switch]{lineno}
\usepackage[utf8]{inputenc} %
\usepackage[T1]{fontenc}    %
\usepackage{hyperref}       %
\usepackage{url}            %
\usepackage{booktabs}       %
\usepackage{amsfonts}       %
\usepackage{nicefrac}       %
\usepackage{microtype}      %
\usepackage[table]{xcolor}         %
\usepackage{enumitem}
\usepackage{tikz}
\usepackage{bm}
\usepackage{amsmath}
\usepackage{tabu}
\usepackage{rotating}
\usepackage{bookmark}
\usepackage{capt-of,etoolbox}
\usepackage{xspace}
\usepackage{xstring}
\usepackage{microtype}
\usepackage{soul}
\usepackage{wrapfig}
\usepackage{graphbox}
\usepackage{tcolorbox}
\usepackage{makecell}
\usepackage{mathtools}
\usepackage{balance}
\usepackage[varqu]{zi4}
\usepackage[bb=boondox]{mathalfa}
\usepackage{units}
\usepackage{tabularx}
\usepackage{tabularray}
\usepackage{gensymb}
\usepackage{printlen} %
\usepackage{placeins} %
\usepackage{afterpage} %
\usepackage{eqnarray}
\usepackage{cleveref}
\usepackage{float}
\usepackage{array}
\usepackage{graphicx}
\usepackage{xcolor}
\usepackage{multirow}
\usepackage{makecell}
\usepackage{geometry}
\def\etal.{et\penalty50\ al.} %
\definecolor{niceblue}{RGB}{56, 116, 203}

\hypersetup{colorlinks, 
      linkcolor=niceblue,
      citecolor=niceblue,
      urlcolor=niceblue,
      filecolor=niceblue}
\usepackage{todonotes}

\title{RenderBender: A Survey on Adversarial Attacks \\ Using Differentiable Rendering}

\author{
Matthew Hull$^1$\and
Haoran Wang$^1$\and
Matthew Lau$^1$\and
Alec Helbling$^1$\and\\
Mansi Phute$^1$\and
Chao Zhang$^1$\and
Zsolt Kira$^1$\and
Willian Lunardi$^2$\and\\
Martin Andreoni$^2$\and
Wenke Lee$^1$\And
Polo Chau$^1$
\affiliations
$^{1}$Georgia Tech, $^{2}$Technology Innovation Institute\\
\small
\emails
$^{1}$\{matthewhull,haoran.wang, mlau40, alechelbling, mphute6, chaozhang, zkira, wenke, polo\}@gatech.edu,
$^{2}$\{willian.lunardi, martin.andreoni\}@tii.ae
}
\normalsize

\begin{document}

\definecolor{agreen}{RGB}{74, 198, 148}
\definecolor{purple}{RGB}{158, 62, 177}
\definecolor{darkpurple}{RGB}{170, 70, 210}
\definecolor{aqua}{RGB}{87, 180, 181}
\definecolor{lightblue}{RGB}{72, 123, 232}
\definecolor{hotpink}{RGB}{255, 83, 115}
\definecolor{teal}{RGB}{90, 200, 250}
\definecolor{linkColor}{RGB}{0, 128, 229}
\definecolor{lightgreen}{RGB}{33, 222, 128}
\definecolor{almostBlack}{RGB}{60,60,60}

\definecolor{red}{RGB}{236, 107, 44}
\definecolor{green}{RGB}{0, 128, 0}
\definecolor{yellow}{RGB}{255, 192, 0}
\definecolor{purple}{RGB}{128, 0, 128}
\definecolor{cyan}{RGB}{0, 255, 255}
\definecolor{lightgray}{gray}{0.95}
\definecolor{grayborder}{gray}{0.5}
\definecolor{gray}{gray}{0.75}
\definecolor{orange}{RGB}{236, 107, 44}
\definecolor{lightorange}{RGB}{255, 223, 186}
\definecolor{blue}{RGB}{116, 95, 232}
\definecolor{lightblue}{RGB}{255, 223, 186}
\definecolor{lightpurple}{RGB}{202,58,126}

\definecolor{9colorq1}{RGB}{166,206,227}
\definecolor{9colorq2}{RGB}{31,120,180}
\definecolor{9colorq3}{RGB}{178,223,138}
\definecolor{9colorq4}{RGB}{51,160,44}
\definecolor{9colorq5}{RGB}{251,154,153}
\definecolor{9colorq6}{RGB}{227,26,28}
\definecolor{9colorq7}{RGB}{253,191,111}
\definecolor{9colorq8}{RGB}{255,127,0}
\definecolor{9colorq9}{RGB}{202,178,214}

\newcommand{\link}[1]{{\href{#1}{\color{blue}\textbf{\texttt{#1}}}}}
\newcommand{\figpart}[1]{\textcolor{blue}{#1}}

\renewcommand{\figurename}{Fig.}

\newcommand{\tool}{\textsc{\textsf{Revamp}}}
\newcommand{\persona}{Jim}
\newcommand{\diffr}{differentiable rendering\xspace}
\newcommand{\Diffr}{Differentiable rendering\xspace}

\definecolor{codebg}{gray}{0.96}
\definecolor{codeframe}{gray}{0.8}
\definecolor{code_gray}{RGB}{115, 115, 115}
\definecolor{atck_sfc_blue}{RGB}{116, 95, 232}
\definecolor{manipulate_3d_rep_pink}{RGB}{202, 58, 126}
\definecolor{designate_domain_yellow}{RGB}{234, 182, 86}
\newcommand{\inlinecode}[1]{\fcolorbox{codeframe}{codebg}{\small\!\texttt{#1}\!}}

\newtoggle{inheader}
\newcommand{\bluehl}[1]{{\sethlcolor{soulblue}\hl{#1}}}
\newcommand{\bluelighthl}[1]{{\sethlcolor{soulbluelight}\hl{#1}}}
\newcommand{\orangehl}[1]{{\sethlcolor{soulorange}\hl{#1}}}
\newcommand{\orangelighthl}[1]{{\sethlcolor{souldorangelight}\hl{#1}}}

\newcommand{\grayhl}[1]{{\sethlcolor{soulgray}\hl{#1}}}
\newcommand{\graylighthl}[1]{{\sethlcolor{soulgraylight}\hl{#1}}}
\newcommand{\redhl}[1]{{\sethlcolor{soulred}\hl{#1}}}

\newcommand{\inlinefig}[2]{\protect\includegraphics[align=c, height=#1pt]{figs/#2}}

\newcommand{\feature}[1]{\featuretag{\sffamily\small{#1}}}
\newcommand{\ck}{\checkmark}

\newcommand*\myquote[1]{``\textit{#1}''}

\setlength{\fboxsep}{0.9pt} %
\setlength{\fboxrule}{0.7pt} %

\newcommand{\ROUNDRECTSIZE}{0.15}

\newcommand{\roundedrect}[1]{%
\tikz\path[rounded corners=1pt,fill=#1] (0,0) rectangle (\ROUNDRECTSIZE cm,\ROUNDRECTSIZE cm);%
}

\newcommand{\roundedrectb}[2]{%
\tikz\path[rounded corners=1pt,fill=#1, draw=#2] (0,0) rectangle (\ROUNDRECTSIZE cm,\ROUNDRECTSIZE cm);%
}

\newcommand{\roundedrectbhalf}{%
\tikz\path[rounded corners=1pt,draw=black] 
(0,0) rectangle (0.25cm,0.25cm)
(0,0) -- (0.25cm,0.25cm) -- (0.25cm,0) -- cycle [fill=white]
(0,0) -- (0.25cm,0.25cm) -- (0,0.25cm) -- cycle [fill=black];
}

\newcommand{\CIRCRAD}{0.08}

\newcommand{\roundedbox}[1]{%
    \tikz[baseline=(X.base)]\node (X) [draw=grayborder, rounded corners=2pt, minimum height=2ex, inner sep=0.1ex] {\strut #1};%
} 

\newcommand{\circlecustom}[2]{%
\tikz\path[fill=#1, draw=#2] (0,0) circle (\CIRCRAD cm);%
}

\newcommand{\circbw}{\circlecustom{white}{black}}

\newcommand{\circb}{%
\tikz\path[fill=black, draw=black] (0,0) circle (\CIRCRAD cm);%
}

\newcommand{\circg}{%
\tikz\path[fill=gray, draw=gray] (0,0) circle (\CIRCRAD cm);%
}

\newcommand{\circp}{%
\tikz[baseline={(0,-0.1)}]{ %
    \path[draw=black, fill=white] (0,0) circle (\CIRCRAD cm); 
    \path[fill=white, draw=none] (0,0) -- (90:\CIRCRAD cm) arc[start angle=90, end angle=210, radius=\CIRCRAD cm] -- cycle;
    \path[fill=gray, draw=none] (0,0) -- (210:\CIRCRAD cm) arc[start angle=210, end angle=330, radius=\CIRCRAD cm] -- cycle;
    \path[fill=black, draw=none] (0,0) -- (330:\CIRCRAD cm) arc[start angle=330, end angle=450, radius=\CIRCRAD cm] -- cycle;
}}

\newcommand{\circh}{%
\tikz\path[draw=black, fill=white] (0,0) circle (\CIRCRAD cm)
  (0,0) -- (90:\CIRCRAD cm) arc[start angle=90, end angle=-90, radius=\CIRCRAD cm] -- cycle [fill=black];
}

\newcolumntype{R}[1]{>{\raggedleft\arraybackslash}p{#1}}
\newcolumntype{L}[1]{>{\raggedright\arraybackslash}p{#1}}
\newcolumntype{C}[1]{>{\centering\arraybackslash}p{#1}}
\newcommand{\g}{\roundedrect{gray}}

\maketitle

\label{sec:abstract}
\begin{abstract}
    Differentiable rendering techniques like Gaussian Splatting and Neural Radiance Fields have become powerful tools for generating high-fidelity models of 3D objects and scenes. 
    Their ability to produce both physically plausible and differentiable models of scenes are key ingredient needed to produce physically plausible adversarial attacks on DNNs.  
    However, the adversarial machine learning community has yet to fully explore these capabilities, partly due to differing attack goals (e.g., misclassification, misdetection) and a wide range of possible scene manipulations used to achieve them (e.g., alter texture, mesh).
    This survey contributes the first framework that unifies diverse goals and tasks, facilitating easy comparison of existing work, identifying research gaps, and highlighting future directions—ranging from expanding attack goals and tasks to account for new modalities, state-of-the-art models, tools, and pipelines, to underscoring the importance of studying real-world threats in complex scenes.
\end{abstract}
\section{Introduction}
\label{sec:introduction}
\begin{figure*}[t!]
    \centering
    \includegraphics[width=1.0\linewidth]{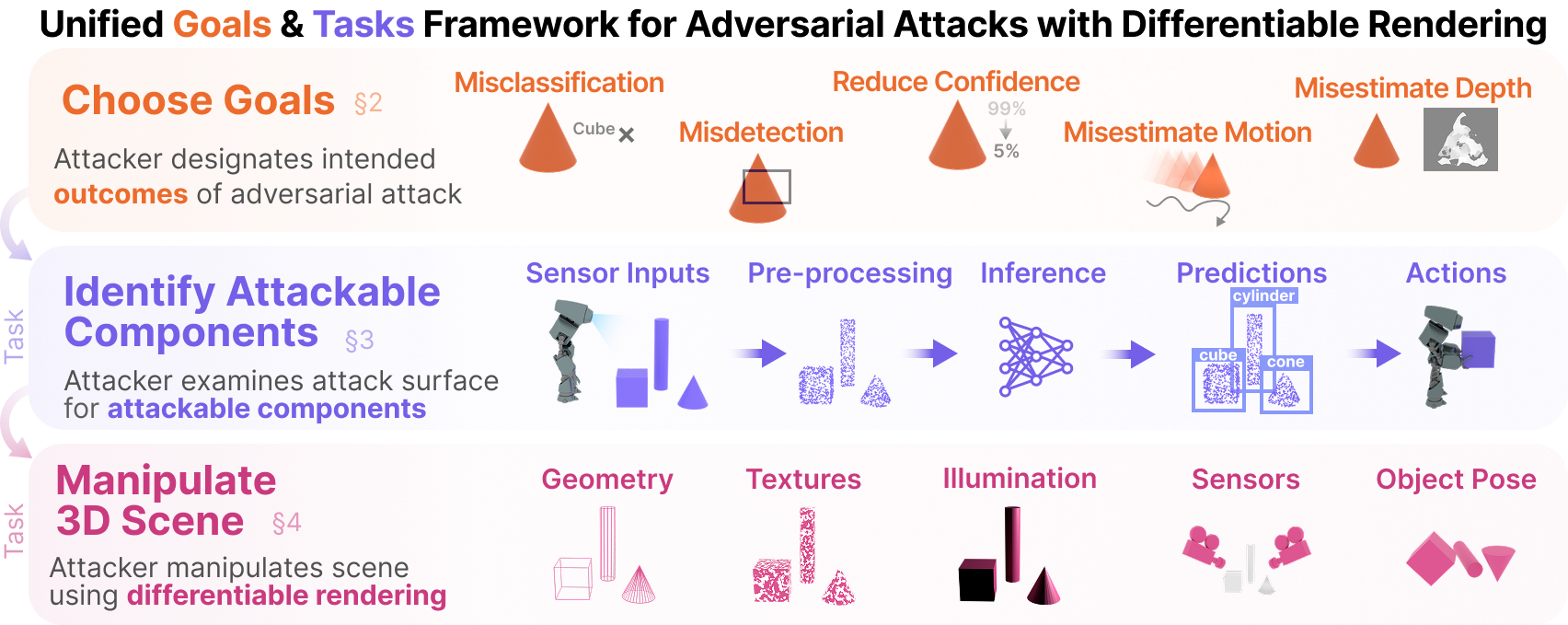}
    \caption{Visual overview of our unifying survey framework that, by unifying the diverse goals and tasks in identifying attackable components and manipulating scene representations, enables systematic summarization and comparison with existing differentiable rendering related adversarial attack research.}
    \label{fig:crown_jewel}
\end{figure*}
Differentiable rendering has emerged as a powerful tool for solving inverse problems in vision and graphics by enabling gradient propagation through the rendering process. 
Recent methods like Neural Radiance Fields (NeRF) \cite{mildenhall_nerf_2020} and 3D Gaussian Splatting \cite{kerbl_3d_2023} enable novel view synthesis from limited images to reconstruct 3D models or scenes. 
These advancements have spurred open-source tools, such as PyTorch3D\footnote{\url{https://pytorch3d.org}} and user-friendly platforms\footnote{\url{https://poly.cam}} that allow creating textured 3D models from photos.

Differentiable rendering has also exposed vulnerabilities in DNNs by enabling adversarial attacks. 
Adversaries exploit DNN gradients to optimize inputs, training, or outputs for malicious purposes, leading to misclassifications in systems such as stop signs in cars, 
LiDAR \cite{cao_adversarial_2019}, facial recognition 
, and 3D models \cite{xiao_meshadv_2019}.
Similarly, \diffr allows attackers to optimize 3D scene parameters (objects, materials, lighting) via loss gradients.
Research on \diffr-based attacks is scattered across:
\begin{enumerate}[topsep=0mm,leftmargin=*,itemsep=0mm]
\item \textbf{Attack goals}: e.g., inducing misclassifications or motion/depth errors;
\item \textbf{Attackable components}: e.g., preprocessing steps or during inference;
\item \textbf{Scene manipulation}: e.g., targeting texture, geometry, or combinations thereof.
\end{enumerate}

In other words, progress in adversarial attacks using \diffr has been made, but systematic comparisons, summaries of strengths, and research gap identification remain challenging. 
Figure \ref{fig:crown_jewel} shows how our survey addresses this gap by organizing tasks like texture manipulation, illumination changes, and 3D mesh alterations, emphasizing both techniques and their potential exploitation by adversaries.

\subsection{Related Survey and Methodology}
\label{subsec:related_surveys_methodology}
This is the first survey to focus on task-based \diffr capabilities for 3D adversarial attacks. Existing work separates \diffr and adversarial research. Kato \etal.~[\citeyear{kato_differentiable_2020}] briefly mention adversarial attacks as an open problem but do not propose a framework distinguishing attacks using detailed goals and tasks.
Since then, NeRF and 3D Gaussian Splatting have gained prominence, requiring discussion. Surveys on NeRF \cite{xie_neural_2022,tewari_advances_2022,gao_nerf_2023,mittal_neural_2024} and 3D Gaussian Splatting \cite{chen_survey_2024,tosi_how_2024} do not address adversarial use. Existing adversarial attack surveys cover 2D/3D models \cite{li_survey_2024}, robustness and defenses \cite{miller_adversarial_2020}, or image classification \cite{machado_adversarial_2023} but omit \diffr.

We reviewed 28 works from top venues in computer vision, ML, and graphics, covering \diffr methods (e.g., NeRF, 3D Gaussian Splatting) and their use in adversarial attacks. 
Using a task-based framework, we categorized attacker goals—texture, illumination, and mesh manipulation—to clarify methodologies and vulnerabilities. As a newer field, \diffr research began in 2014, with adversarial applications emerging in 2019.

\subsection{Contributions}
\begin{enumerate}[itemindent=*, leftmargin=0pt, label=\textbf{C\arabic*.}, ref=\textbf{C\arabic*}]
    \item \textbf{We present the first comprehensive, attacker-task guided survey on adversarial attacks using \diffr, incorporating a use-inspired approach} (Fig. \ref{fig:crown_jewel}).
    Our framework positions each work by attacker objectives and \diffr techniques, defining the attack surface based on feasible scene manipulations of the scene representations (Sec. \ref{subsec:attack-surface}).
    \begin{itemize}[leftmargin=*,itemsep=0pt]
    \item Our methodology links goals to tasks, providing a structured comparison of works and identifying research gaps.
    \item Table \ref{tab:attacker-goals-tasks-overview} explains \diffr's role in attacks, relevant methods, and current strengths and limitations.
    \end{itemize}
    \item \textbf{We provide comprehensive categorizations of attack methods, highlighting their impact and real-world implications} (Sec. \ref{sec:identify-attackable-comps}).
    We show a ``Target List'' of attacked models, including object detection, image classification, and others along with attacker access levels (Table \ref{tab:overview-dnn-models-attacked}).
    These resources enable researchers to build on existing work, compare outcomes, and develop new techniques to address adversarial threats using \diffr.
    \item \textbf{We identify key future research directions to address the growing threat of adversarial attacks} (Sec. \ref{sec:research-directions}).
    Priorities include developing robust defenses, exploring novel attack strategies, and investigating the physical plausibility of these attacks.
\end{enumerate}

\section{Attacker Goals}
\label{sec:attacker-goals}
To better understand how differentiable rendering is used in adversarial attacks, we describe an attacker using the threat model concept to delineate their goals, capabilities, and knowledge in the context of the attack they wish to carry out \cite{li_survey_2024}.
Using an attacker-task guided perspective, we connect the attacker goals to required tasks and sub-tasks (Sec. \ref{sec:identify-attackable-comps}) that are used in differentiable rendering attacks.
Attacker goals encompass any threat affecting the integrity of a DNN's intended task \cite{papernot_towards_2016,li_survey_2024,wiyatno_adversarial_2019}.
In this survey, we identify five attacker goals that are used in attacks on deep learning models using \diffr:
\begin{enumerate}[itemindent=*, itemsep=0pt, leftmargin=0pt, label=\textbf{2.\arabic*.}, ref=\textbf{2.\arabic*}]
\item \textbf{Misclassification} - the model predicts an incorrect class (untargeted) or a specified incorrect class (targeted) \cite{papernot_limitations_2016}.
\label{subsec:attacker-goals-misclassification}
\item \textbf{Misdetection} - manifested as various errors: nothing is detected (evasion), improper bounding box localization, duplicate detections, or detecting background as an object, or combinations thereof \cite{bolya_tide_2020}. 
\label{subsec:attacker-goals-misdetection}
\item \textbf{Reduce Confidence} - the target class is not predicted with high confidence \cite{papernot_limitations_2016}. 
\label{subsec:attacker-goals-reduce-confidence}
\item \textbf{Misestimate Motion} - the model misestimates the motion of objects in the scene caused by adversarial movements or objects \cite{schmalfuss_distracting_2023}.
\label{subsec:attacker-goals-misestimate-motion}
\item \textbf{Misestimate Depth} - the model misestimates depth, affecting the model's ability to perceive distances. \cite{zheng_physical_2024}.
\label{subsec:attacker-goals-misestimate-depth}
\end{enumerate} 
In Table \ref{tab:attacker-goals-tasks-overview}, we categorized our 28 survey papers as S=Survey (3), M=Metrics (1), or A=Attack (24).  
Of our 24 Attack papers, we found that 18 works chose goals of inducing misclassifications, 17 induced misdetections, and 10 induced reduction in model confidence while only 1 work each pursued attack goals of misestimation of motion or depth.

\newlength{\colwidth}
\setlength{\colwidth}{0.6cm}

\begin{table*}[t!]
    \centering
    \sffamily
    \tiny
    \arrayrulecolor{black}
    \setlength{\arrayrulewidth}{1pt}
    \renewcommand{\arraystretch}{0.7}
    \rowcolors{2}{lightgray}{white}
    \setlength{\tabcolsep}{1pt}
    \begin{tabular}{r !{\color{black}\vrule width 0.5pt} C{\colwidth} C{\colwidth} C{\colwidth} C{\colwidth} C{\colwidth} 
        !{\color{black}\vrule width 0.5pt} C{\colwidth} C{\colwidth}
        !{\color{grayborder}\vrule width 0.25pt} C{\colwidth} C{\colwidth} C{\colwidth} C{\colwidth} C{\colwidth} 
        !{\color{grayborder}\vrule width 0.25pt} C{\colwidth} C{\colwidth} 
        !{\color{black}\vrule width 0.5pt} C{\colwidth}
        !{\color{black}\vrule width 0.5pt} C{\colwidth} C{\colwidth}}
    \multicolumn{1}{c}{} & 
    \multicolumn{5}{c}{\textcolor{orange}{\makecell{\textbf{ATTACKER GOALS}}}} & 
    \multicolumn{7}{c}{\textcolor{blue}{\makecell{\cellcolor{white}\textbf{REQUIRED TASKS} }}} & 
    \multicolumn{2}{c}{\textcolor{orange}{\makecell{\cellcolor{white}\textbf{DOMAIN} }}} & 
    \multicolumn{1}{c}{\textcolor{grayborder}{\textbf{}}} & 
    \multicolumn{2}{c}{\textcolor{grayborder}{\textbf{WHERE}}} \\        
    \cmidrule(lr){2-6} \cmidrule(lr){7-13} \cmidrule(lr){14-15} \cmidrule(lr){17-18}
    \multicolumn{1}{c}{\cellcolor{white}{}} & 
    \multicolumn{5}{c}{\textcolor{orange}{\cellcolor{white}{}}} & 
    \multicolumn{2}{c}{\textcolor{atck_sfc_blue}{\makecell{\cellcolor{white}\textbf{\tiny{IDENTIFY}}}}} & 
    \multicolumn{5}{c}{\textcolor{manipulate_3d_rep_pink}{\makecell{\cellcolor{white}{\textbf{\tiny{MANIPULATE}}}}}} & 
    \multicolumn{2}{c}{} & 
    \multicolumn{2}{c}{\textcolor{grayborder}{}}\\
    \rowcolor{white}
    \multicolumn{1}{r}{\textbf{Work \quad}} & 
    \rotatebox{90}{\roundedbox{2.1} Misclassification} & 
    \rotatebox{90}{\roundedbox{2.2} Missed Detection} & 
    \rotatebox{90}{\roundedbox{2.3} Reduce Model Confidence} & 
    \rotatebox{90}{\roundedbox{2.4} Misestimation of Motion} & 
    \rotatebox{90}{\roundedbox{2.5} Misestimation of Depth} & 
    \rotatebox{90}{\roundedbox{3.1} Attack Surface} & 
    \rotatebox{90}{\roundedbox{3.2} Analyze Scene Components} & 
    \rotatebox{90}{\roundedbox{4.1} Attack Scene Geometry} & 
    \rotatebox{90}{\roundedbox{4.2} Attack Scene Textures} & 
    \rotatebox{90}{\roundedbox{4.3} Attack Scene Object Pose} & 
    \rotatebox{90}{\roundedbox{4.4} Attack Scene Illumination} & 
    \rotatebox{90}{\roundedbox{4.5} Attack Scene Sensors} & 
    \rotatebox{90}{\roundedbox{5.1} Perform Digital Attack} & 
    \rotatebox{90}{\roundedbox{5.2} Perform Physical Attack} & 
    \rotatebox{90}{Paper Type} & 
    \multicolumn{2}{c}{\rotatebox{90}{Publication Venue} } \\ 
    \midrule 
    \cite{abdelfattah_towards_2021}     &    & \g &    &    &    &    &    & \g & \g &    &    &    & \g &    & A  & \multicolumn{2}{l}{ICIP} \\
    \cite{alcorn_strike_2019}           & \g & \g &    &    &    &    &    &    &    & \g &    &    & \g &    & A  & \multicolumn{2}{l}{CVPR} \\
    \cite{bolya_tide_2020}              & \g & \g &    &    &    &    &    &    &    &    &    &    & \g &    & M  & \multicolumn{2}{l}{ECCV}\\ 
    \cite{byun_improving_2022}          & \g &    &    &    &    &    &    &    & \g &    &    &    & \g &    & A  & \multicolumn{2}{l}{arXiv} \\
    \cite{cao_adversarial_2019}         & \g & \g &    &    &    &    &    & \g &    &    &    &    & \g &    & A  & \multicolumn{2}{l}{arXiv} \\
    \cite{dong_viewfool_2022}           & \g &    &    &    &    &    &    &    &    & \g &    &    & \g & \g & A  & \multicolumn{2}{l}{NeurIPS} \\
    \cite{huang_towards_2024}           & \g & \g &    &    &    &    &    & \g & \g &    &    &    & \g & \g & A  & \multicolumn{2}{l}{CVPR} \\
    \cite{li_survey_2024}               & \g & \g & \g &    &    & \g & \g &    &    &    &    &    & \g &    & S  & \multicolumn{2}{l}{ACM CSUR} \\ 
    \cite{li_adv3d_2023}                &    & \g & \g &    &    &    &    &    & \g &    &    &    & \g &    & A  & \multicolumn{2}{l}{arXiv} \\
    \cite{li_flexible_2024}             & \g & \g &    &    &    &    &    &    & \g &    &    &    & \g & \g & A  & \multicolumn{2}{l}{arXiv} \\
    \cite{liu_beyond_2019}              & \g & \g &    &    &    &    &    & \g & \g &    & \g &    & \g &    & A  & \multicolumn{2}{l}{ICLR} \\
    \cite{machado_adversarial_2023}     & \g &    & \g &    &    &    &    &    &    &    &    &    & \g &    & S  & \multicolumn{2}{l}{ACM}\\ 
    \cite{maesumi_learning_2021}        &    & \g & \g &    &    &    &    &    & \g &    &    &    & \g &    & A  & \multicolumn{2}{l}{arXiv} \\
    \cite{meloni_messing_2021}          & \g &    & \g &    &    &    &    &    & \g &    &    &    & \g &    & A  & \multicolumn{2}{l}{ICMLA} \\
    \cite{papernot_towards_2016}        & \g & \g & \g &    &    & \g & \g &    &    &    &    &    & \g &    & A  & \multicolumn{2}{l}{EuroS\&P} \\ 
    \cite{papernot_limitations_2016}    & \g & \g & \g &    &    & \g & \g &    &    &    &    &    & \g &    & A  & \multicolumn{2}{l}{EuroS\&P} \\ 
    \cite{schmalfuss_distracting_2023}  &    &    &    & \g &    &    &    & \g & \g & \g &    &    & \g &    & A  & \multicolumn{2}{l}{ICCV} \\
    \cite{shahreza_comprehensive_2023}  & \g &    &    &    &    & \g & \g &    &    & \g &    & \g & \g & \g & A  & \multicolumn{2}{l}{TPAMI} \\
    \cite{suryanto_dta_2022}            & \g & \g &    &    &    &    &    &    & \g &    &    &    & \g &    & A  & \multicolumn{2}{l}{CVPR} \\
    \cite{suryanto_active_2023}         & \g & \g & \g &    &    &    &    &    & \g &    &    &    & \g & \g & A  & \multicolumn{2}{l}{ICCV} \\
    \cite{tu_exploring_2021}            &    & \g & \g &    &    &    &    & \g & \g &    &    &    & \g &    & A  & \multicolumn{2}{l}{CoRL} \\
    \cite{wang_fca_2022}                & \g & \g & \g &    &    &    &    &    & \g &    &    &    & \g & \g & A  & \multicolumn{2}{l}{AAAI} \\
    \cite{wiyatno_adversarial_2019}     & \g & \g & \g &    &    & \g & \g &    &    &    &    &    & \g &    & A  & \multicolumn{2}{l}{arXiv} \\
    \cite{xiao_meshadv_2019}            & \g & \g &    &    &    &    &    & \g &    &    &    &    & \g &    & A  & \multicolumn{2}{l}{CVPR} \\
    \cite{yuan_adversarial_2019}        & \g & \g & \g &    &    & \g & \g &    &    &    &    &    & \g &    & S  & \multicolumn{2}{l}{TNNLS} \\ 
    \cite{zeng_adversarial_2019}        & \g &    &    &    &    &    &    &    &    & \g & \g &    & \g & \g & A  & \multicolumn{2}{l}{CVPR} \\
    \cite{zheng_physical_2024}          &    &    &    &    & \g &    &    &    & \g &    &    &    & \g & \g & A  & \multicolumn{2}{l}{CVPR} \\
    \cite{zhou_rauca_2024}              & \g & \g & \g &    &    &    &    &    & \g &    &    &    & \g & \g & A  & \multicolumn{2}{l}{ICML} \\
    \bottomrule
\end{tabular}
\rmfamily
\normalsize
\centering

\caption{Overview of representative works on adversarial attacks using differentiable rendering methods.
Each row is one work; each column corresponds to a required attacker task or goal. A work's relevant goal or task is indicated by a colored cell.
S = Survey, M = Metrics, A = Attack.}
\label{tab:attacker-goals-tasks-overview}
\end{table*}

\section{Identify Attackable Components}
\label{sec:identify-attackable-comps}
To achieve the attacker goals in Sec. \ref{sec:attacker-goals}, one must identify which components can be manipulated by analyzing the attack surface (Sec. \ref{subsec:attack-surface}) and understanding 3D scene representations.

\subsubsection{Attack Surface}
\label{subsec:attack-surface}
The attack surface includes all data processing stages \cite{papernot_towards_2016} in Fig. \ref{fig:crown_jewel}, from sensor inputs and pre-processing to model inference and output actions. In \diffr, this surface extends to the renderer and scene representation, giving adversaries multiple potential entry points. For instance, a robot scanning its 3D environment has:
\begin{itemize}[itemindent=*, leftmargin=0pt, topsep=0mm, itemsep=0mm]
\item \textbf{Sensor Inputs} (e.g., camera, LiDAR).
\item \textbf{Pre-processing} steps (e.g., generating 2D images or point clouds).
\item \textbf{Inference} by the DNN model.
\item \textbf{Predictions} (e.g., labels, bounding boxes, segmentation).
\item \textbf{Actions} or decisions based on model output.
\end{itemize}
An attack’s effectiveness hinges on the adversary’s \textbf{access level}: white-box \circbw, black-box \circb, or combination thereof \circh, classified in Table \ref{tab:overview-dnn-models-attacked}. 
In \diffr, attackers manipulate scene elements (Sec. \ref{subsec:understand-3d-representations}), such as object textures or environmental factors (e.g., adversarial weather \cite{schmalfuss_distracting_2023}) to deceive the DNN.
\subsubsection{Analyze Scene Components}
\label{subsec:understand-3d-representations}
In \diffr attacks, the 3D \textit{scene representation} 
is the main target. We categorize its components under:
\textbf{Geometry} (explicit or implicit \cite{mildenhall_nerf_2020,kerbl_3d_2023}),
\textbf{Texture} (color and reflectance),
\textbf{Position/Pose} (object location/orientation),
\textbf{Illumination} (light sources, e.g., sun or lamps), and
\textbf{Sensors} (camera/LiDAR properties like resolution or field of view).
Identifying these components helps attackers craft manipulations that produce realistic, adversarial inputs to DNN models.
\begin{table}
    \sffamily
    \tiny
    \begin{tblr}{
      rowsep=0.0pt,
      colsep=0.38pt,
      row{even} = {lightgray, c},      row{3,5,7,9,11,13,15,17,19,21,23,25,27,29,31,33,35,37,39,41,43,45,47,49,51,53,55,57,59,61,63,64} = {c},
      row{2,22,24,44,46,52,56,61,63} = {almostBlack, c, fg=white},
      column{1} = {white, c},
      column{2} = {r},
      cell{1}{1} = {c},
      cell{1}{2} = {r},
      cell{2}{1} = {r=21}{},
      cell{23}{1} = {r=0}{},
      cell{24}{1} = {r=22}{},
      cell{46}{1} = {r=0}{},
      cell{47}{1} = {r=6}{},
      cell{53}{1} = {r=6}{},
      cell{59}{1} = {r=3}{},
      vline{3} = {1-73}{},
    }    
     & \textbf{Model \quad} & 
     \rotatebox{90}{\cite{abdelfattah_towards_2021}} & 
     \rotatebox{90}{\cite{alcorn_strike_2019}} & 
     \rotatebox{90}{\cite{byun_improving_2022}} & 
     \rotatebox{90}{\cite{dong_viewfool_2022}} & 
     \rotatebox{90}{\cite{hu_physically_2023}} & 
     \rotatebox{90}{\cite{huang_towards_2024}} & 
     \rotatebox{90}{\cite{jiang_nerfail_2024}} & 
     \rotatebox{90}{\cite{li_adv3d_2023}} & 
     \rotatebox{90}{\cite{li_flexible_2024}} & 
     \rotatebox{90}{\cite{liu_beyond_2019}} & 
     \rotatebox{90}{\cite{maesumi_learning_2021}} & 
     \rotatebox{90}{\cite{meloni_messing_2021}} & 
     \rotatebox{90}{\cite{schmalfuss_distracting_2023}} & 
     \rotatebox{90}{\cite{shahreza_comprehensive_2023}} & 
     \rotatebox{90}{\cite{suryanto_dta_2022}} & 
     \rotatebox{90}{\cite{suryanto_active_2023}} & 
     \rotatebox{90}{\cite{tu_exploring_2021}} & 
     \rotatebox{90}{\cite{wang_fca_2022}} & 
     \rotatebox{90}{\cite{xiao_meshadv_2019}} & 
     \rotatebox{90}{\cite{zeng_adversarial_2019}} & 
     \rotatebox{90}{\cite{zheng_physical_2024}} & 
     \rotatebox{90}{\cite{zhou_rauca_2024}} \\
     & \textbf{Image} &  &  &  &  &  &  &  &  &  &  &  &  &  &  &  &  &  &  &  &  &  & \\
     & {ResNet-18} &  &  & \circb &  &  &  &  &  &  &  &  &  &  &  &  &  &  &  &  &  &  & \\
     & ResNet-34 &  &  &  &  &  &  &  &  &  &  &  &  &  &  &  &  &  &  &  & \circbw &  & \\
     & ResNet-50 &  & \circb & \circh & \circbw &  & \circb & \circbw &  &  &  &  &  &  &  &  &  &  &  &  &  &  & \\
     & ResNet-101 &  &  &  &  &  & \circh &  &  &  & \circbw &  &  &  &  &  &  &  &  &  &  &  & \\
     & ResNet-152 &  &  &  &  &  & \circb &  &  &  &  &  &  &  &  &  &  &  &  &  &  &  & \\
     & AlexNet &  & \circb &  &  &  &  & \circbw &  &  & \circb &  &  &  &  &  &  &  &  &  & \circbw &  & \\
     & VGG-16 &  &  & \circh & \circb &  & \circb & \circbw &  &  & \circb &  &  &  &  &  &  &  &  &  &  &  & \\
     & VGG-19 &  &  &  &  &  & \circb &  &  &  &  &  &  &  &  &  &  &  &  &  &  &  & \\
     & SqueezeNet &  &  &  &  &  &  &  &  &  & \circb &  &  &  &  &  &  &  &  &  &  &  & \\
     & DenseNet  &  &  &  &  &  &  &  &  &  & \circb &  &  &  &  &  &  &  &  & \circbw &  &  & \\
     & DenseNet-121  &  &  & \circh & \circb &  & \circh &  &  &  &  &  &  &  &  &  &  &  &  &  &  &  & \\
     & Inception &  & \circbw & \circh & \circb &  & \circb &  &  &  &  &  & \circbw &  &  &  &  &  &  & \circbw &  &  & \\
     & Inc-ResNet &  &  & \circb & \circb &  &  &  &  &  &  &  &  &  &  &  &  &  &  &  &  &  & \\
     & EfficientNet &  &  &  &  &  & \circb & \circbw &  &  &  &  &  &  &  &  &  &  &  &  &  &  & \\
     & MobileNet-v2 &  &  & \circb & \circb &  & \circb & \circbw &  &  &  &  & \circbw &  &  &  &  &  &  &  &  &  & \\
     & ViT-B/16 &  &  &  & \circbw &  & \circb & \circbw &  &  &  &  &  &  &  &  &  &  &  &  &  &  & \\
     & DeiT-B &  &  &  & \circb &  &  &  &  &  &  &  &  &  &  &  &  &  &  & \circbw &  &  & \\
     & Swin-B &  &  &  & \circb &  & \circb &  &  &  &  &  &  &  &  &  &  &  &  &  &  &  & \\
     & Mixer-B &  &  &  & \circb &  &  &  &  &  &  &  &  &  &  &  &  &  &  &  &  &  & \\
     & \textbf{Semantic Seg.} &  &  &  &  &  &  &  &  &  &  &  &  &  &  &  &  &  &  &  &  &  & \\
     & Mask-RCNN &  &  &  &  & \circb &  &  &  & \circb &  &  &  &  &  & \circb & \circb &  & \circb &  &  &  & \\
     & \textbf{Object Det.} &  &  &  &  &  &  &  &  &  &  &  &  &  &  &  &  &  &  &  &  &  & \\
     & YOLO-v2 &  &  &  &  & \circb &  &  &  &  &  & \circh &  &  &  &  &  &  &  &  &  &  & \\
     & YOLO-v3 & \circbw & \circb &  &  & \circbw &  &  &  & \circbw &  &  &  &  &  &  & \circbw &  & \circh &  &  &  & \circbw \\
     & YOLO-v4 &  &  &  &  &  &  &  &  &  &  &  &  &  &  & \circbw &  &  &  &  &  &  & \\
     & YOLO-v5 &  &  &  &  &  &  &  &  & \circbw &  &  &  &  &  &  &  &  &  &  &  &  & \\
     & YOLO-v7 &  &  &  &  &  &  &  &  & \circb &  &  &  &  &  &  &  &  &  &  &  &  & \\
     & YOLO-X &  &  &  &  &  &  &  &  &  &  &  &  &  &  &  &  &  &  &  &  &  & \circb\\
     & EfficientDet &  &  &  &  &  &  &  &  &  &  &  &  &  &  & \circbw &  &  &  &  &  &  & \\
     & Faster-RCNN &  &  &  &  & \circbw &  &  &  & \circbw &  & \circbw &  &  &  & \circb & \circb &  & \circb &  &  &  & \circb\\
     & Dynamic-RCNN  &  &  &  &  &  &  &  &  &  &  &  &  &  &  &  &  &  &  &  &  &  & \circb\\
     & Sparse-RCNN  &  &  &  &  &  &  &  &  &  &  &  &  &  &  &  &  &  &  &  &  &  & \circb\\
     & Cascade-RCNN &  &  &  &  &  &  &  &  & \circb &  &  &  &  &  &  &  &  &  &  &  &  & \\
     & DETR &  &  &  &  & \circbw &  &  &  & \circb &  &  &  &  &  &  &  &  &  &  &  & \circb & \circb \\  
     & SSD &  &  &  &  &  &  &  &  & \circb &  &  &  &  &  & \circb & \circb &  & \circb &  &  &  & \\
     & PVT &  &  &  &  &  &  &  &  &  &  &  &  &  &  &  &  &  &  &  &  &  & \\
     & FCOS3D &  &  &  &  &  &  &  & \circh &  &  &  &  &  &  &  &  &  &  &  &  &  & \\
     & PGD-DET &  &  &  &  &  &  &  & \circh &  &  &  &  &  &  &  &  &  &  &  &  &  & \\
     & DETR3D &  &  &  &  &  &  &  & \circh &  &  &  &  &  &  &  &  &  &  &  &  &  & \\
     & BEV-DET &  &  &  &  &  &  &  & \circh &  &  &  &  &  &  &  &  &  &  &  &  &  & \\
     & Grounding DINO &  &  &  &  &  & \circb &  &  &  &  &  &  &  &  &  &  &  &  &  &  &  & \\
     & \textbf{Point Cloud} &  &  &  &  &  &  &  &  &  &  &  &  &  &  &  &  &  &  &  &  &  & \\
     & Fr-PointNet & \circbw &  &  &  &  &  &  &  &  &  &  &  &  &  &  &  &  &  &  &  &  & \\
     & \textbf{Opt. Flow Est.} &  &  &  &  &  &  &  &  &  &  &  &  &  &  &  &  &  &  &  &  &  & \\
     & FlowNet &  &  &  &  &  &  &  &  &  &  &  &  & \circbw &  &  &  &  &  &  &  &  & \\
     & SpyNet &  &  &  &  &  &  &  &  &  &  &  &  & \circbw &  &  &  &  &  &  &  &  & \\
     & RAFT &  &  &  &  &  &  &  &  &  &  &  &  & \circbw &  &  &  &  &  &  &  &  & \\
     & GMA &  &  &  &  &  &  &  &  &  &  &  &  & \circbw &  &  &  &  &  &  &  &  & \\
     & FFormer &  &  &  &  &  &  &  &  &  &  &  &  & \circbw &  &  &  &  &  &  &  &  & \\
     & \textbf{Face Recog.} &  &  &  &  &  &  &  &  &  &  &  &  &  &  &  &  &  &  &  &  &  & \\
     & ArcFace &  &  &  &  &  &  &  &  &  &  &  &  &  & \circbw &  &  &  &  &  &  &  & \\
     & Elastiface &  &  &  &  &  &  &  &  &  &  &  &  &  & \circbw &  &  &  &  &  &  &  & \\
     & FaceX-Zoo &  &  &  &  &  &  &  &  &  &  &  &  &  & \circb &  &  &  &  &  &  &  & \\     
     & \textbf{Depth Est.} &  &  &  &  &  &  &  &  &  &  &  &  &  &  &  &  &  &  &  &  &  & \\
     & Monodepth2 &  &  &  &  &  &  &  &  &  &  &  &  &  &  &  &  &  &  &  &  & \circbw & \\
     & Depthhints &  &  &  &  &  &  &  &  &  &  &  &  &  &  &  &  &  &  &  &  & \circbw & \\
     & Manydepth &  &  &  &  &  &  &  &  &  &  &  &  &  &  &  &  &  &  &  &  & \circbw & \\
     & Robustdepth &  &  &  &  &  &  &  &  &  &  &  &  &  &  &  &  &  &  &  &  & \circbw & \\
     & \textbf{Fused} &  &  &  &  &  &  &  &  &  &  &  &  &  &  &  &  &  &  &  &  &  & \\
     & MMF &  &  &  &  &  &  &  &  &  &  &  &  &  &  &  &  & \circbw &  &  &  &  & \\
     & \textbf{VLP} &  &  &  &  &  &  &  &  &  &  &  &  &  &  &  &  &  &  &  &  &  & \\
     & BLIP &  &  &  &  &  & \circbw &  &  &  &  &  &  &  &  &  &  &  &  &  &  &  & 
    \end{tblr}
    \vspace{-0.5em}
    \caption{\small{DNNs attacked by differentiable rendering. Each column is one work; each row is a  model.}}
    \label{tab:overview-dnn-models-attacked}    
    \end{table}
    \normalsize
\section{Manipulate 3D Scene}
\label{sec:manipulations}
 
\Diffr enables gradient-based manipulation of any scene elements. 
With white-box access to victim models, attackers can use loss gradients with respect to scene representations to guide such manipulations. 
This section reviews common manipulations on 3D scene representations, including geometry, texture, pose, illumination, and sensors. 
Among the surveyed works, texture attacks are the most prevalent (15) since first adversarial works were on 2D images, followed by geometry (7), pose (5), illumination (2), and sensors (1). 
\vspace{1cm}
\subsection{Attacks on Scene Geometry}
\label{subsec:attack-scene-geometry}
\begin{wrapfigure}{l}{0.05\textwidth}
    \vspace{-15pt} 
    \centering
    \includegraphics[width=0.09\textwidth]{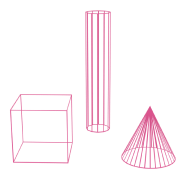}
    \vspace{-25pt} 
\end{wrapfigure}
\textbf{Mesh.} Attackers use \diffr to generate adversarial meshes by perturbing vertex positions to minimize the cross-entropy loss towards the target label. The adversarial meshes are re-rendered as inputs to victim models. 
Beyond Pixel Norm-Balls \cite{liu_beyond_2019} introduced a differentiable rendering framework for generating adversarial geometry $V^\prime$ by propagating gradients through a rendering pipeline via chain rule: 
\begin{equation}
V^\prime \leftarrow V - \gamma \dfrac{\partial C}{\partial I} \dfrac{\partial I}{\partial N} \dfrac{\partial N}{\partial V},
\label{eq:chain_rule}
\end{equation}
where $V$ are vertex positions, $N$ per-face normals, and $\gamma$ the attack strength.
MeshAdv \cite{xiao_meshadv_2019} used Neural Mesh Renderer 
to perturb vertices, attacking classifiers and object detectors like YOLO-v3. TT3D \cite{huang_towards_2024} created adversarial geometry via NeRF and marching cubes but faced scalability challenges due to optimization overhead \cite{tewari_advances_2022}.
Distracting Downpour \cite{schmalfuss_distracting_2023} attacked optical flow models by adding scene-specific spatiotemporally consistent particulate geometry (e.g., rain or snow) to create false motion signals in various datasets.
 
\textbf{Point Cloud.} LiDAR-ADV \cite{cao_adversarial_2019} used a differentiable LiDAR simulator to perturb point clouds, converting the initially non-differentiable features into differentiable ones with smoothing. 
Two other works perturbed point cloud objects and converted them to textured meshes to target multi-modal systems \cite{abdelfattah_towards_2021,tu_exploring_2021}. 

\textbf{Geometry Post-Processing and Stabilization.} Post-perturbation processing maintains realism and avoids topological issues, such as self-intersections or non-manifold meshes. 
Techniques like Laplacian smoothing 
, regularization loss 
, and Chamfer distance loss 
ensure realistic and stable adversarial geometry. 
For instance, Laplacian smoothing minimizes deviations between original and adversarial vertex,
while Chamfer distance loss penalizes dissimilarities between point clouds $P$ and $Q$.
Depth completion and lighting approximation from Tu et al.~[\citeyear{tu_exploring_2021}] enhance realism by restricting adversary scale within axis-aligned bounding boxes. 
\vspace{-0.2cm}
\subsection{Attack Scene Texture}
\label{subsec:attack-scene-texture}
\begin{wrapfigure}{l}{0.05\textwidth}
    \vspace{-15pt} 
    \centering
    \includegraphics[width=0.08\textwidth]{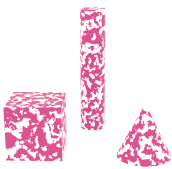}
    \vspace{-22pt} 
\end{wrapfigure}
Texture adversarial attacks manipulate an object's appearance by perturbing its color, pattern, or light reflection properties. 
Using \diffr, the model's loss gradient is used to perturb the texture mappings (e.g., UV maps) via world-aligned methods that optimize 2D textures, UV map-based methods that directly optimize 3D textures, and neural-rendered methods that dynamically generate textures from 3D representations \cite{zhou_rauca_2024}.

\textbf{Multi-Object Texture Attacks.} Two works explore multi-object texture attacks to study transferability. 
Meloni et al.~[\citeyear{meloni_messing_2021}] create poisoned data by perturbing texels using a saliency map from a non-differentiable renderer. 
Byun et al.~[\citeyear{byun_improving_2022}] demonstrate transferability by applying 2D adversarial textures to various 3D objects, achieving successful impersonation and dodging attacks against facial recognition classifiers. 

\textbf{Adversarial Camouflage.} Adversarial camouflage targets vehicles and humans. 
For vehicles, FCA \cite{wang_fca_2022} applied adversarial textures to an Audi e-Tron in CARLA scenes using the Neural Mesh Renderer (NMR) 
, while DTA \cite{suryanto_dta_2022} used EoT for texture projections for a Tesla Model 3 and ACTIVE \cite{suryanto_active_2023} made a further improvement with tri-planar mapping, allowing complex shapes. 
Li et al.~[\citeyear{li_flexible_2024}] conducted a flexible physical camouflage attack (FPA) using diffusion models to generate UV-map-based textures, improving the environmental adaptability in neural rendering. 
RAUCA \cite{zhou_rauca_2024} extended this by incorporating environmental conditions via an encoder-decoder Environmental Feature Extractor (EFE) for optimized textures. 
For humans, Maesumi et al.~[\citeyear{maesumi_learning_2021}] developed adversarial clothing using UV maps and SMPL models, using Blender’s subdivision surface modifier to improve texture resolution for more effective attacks.

\textbf{Texture Attacks on Autonomous Driving Systems.} 
Abdelfattah et al. attacked object textures by treating vertex colors as learnable parameters, reducing YOLOv3 detection in cascaded models used in self-driving \cite{tu_exploring_2021}. 
Adv3D \cite{li_adv3d_2023} used NeRF with semantic branch augmentation along with EoT to enhance physical transferability and reduce the confidence of LiDAR detector, while $\mathrm{3D}^2\mathrm{Fool}$ \cite{zheng_physical_2024} developed object-agnostic adversarial patches via EoT and texture conversion to attack monocular depth estimation models.  

\textbf{Texture Post-Processing and Stabilization.} To enhance the appearance and physical transferability of adversarial textures, many works incorporate post-processing techniques such as hyperparameter tuning, Total Variation (TV) loss, Smooth loss, and Non-Printability Score (NPS). 
TV loss \cite{mahendran_understanding_2015} penalizes differences between adjacent texture pixels $\bm{x}$, reducing noise and promoting smoothness:
\begin{equation*}
    TV(\bm{x}) = \sum_{i,j}\big( (\bm{x}_{i,j+1} - \bm{x}_{ij})^2 + (\bm{x}_{i+1,j} - \bm{x}_{ij})^2 \big)^{\frac{1}{2}}.
\end{equation*}
NPS \cite{sharif_accessorize_2016} assesses the physical printability of textures by evaluating pixel proximity to printable RGB triplets $P \subset [0,1]^3$:
\begin{equation*}
    NPS(\hat{p}) = \prod_{p \in P}|\hat{p} - p|,
\end{equation*}
where a low score indicates higher printability. These methods ensure adversarial textures are both visually plausible and physically realizable.  
\subsection{Attack Scene Illumination}
\label{subsec:attack-scene-illumination}
\begin{wrapfigure}{l}{0.05\textwidth}
    \centering
    \vspace{-15pt} 
    \includegraphics[width=0.08\textwidth]{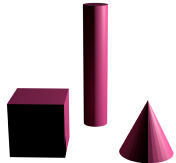}
    \vspace{-25pt} 
\end{wrapfigure}  
Illumination manipulation in \diffr attacks is underexplored due to lack of tools that support controlling light across the whole scene in realistic ways, but two works demonstrate its potential.
Liu et al.~[\citeyear{liu_beyond_2019}] used spherical harmonic lighting 
for global adjustments,
optimizing coefficients via the chain rule (Eq. \ref{eq:chain_rule}) to preserve realism while attacking DNNs. 
Zeng et al.~[\citeyear{zeng_adversarial_2019}] manipulated point lights, creating adversarial lighting to mislead DNNs.
\vspace{-0.2cm}
\subsection{Attack Scene Sensors}
\label{subsec:attack-scene-sensors}
\begin{wrapfigure}{l}{0.05\textwidth}
    \centering
    \vspace{-15pt} 
    \includegraphics[width=0.08\textwidth]{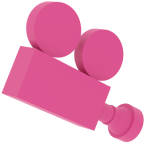}
    \vspace{-25pt} 
\end{wrapfigure}
While many attacks test robustness to camera angle changes, Shahreza et al.~[\citeyear{shahreza_comprehensive_2023}] directly manipulated camera parameters for attacks. 
Using NeRF, they optimized camera rotations to find face poses capable of impersonating target identities in facial recognition models. Their method reconstructs 3D faces from 2D facial templates, enabling practical presentation attacks, such as digital screen replay or printed photographs, which can be further extended to create wearable face masks for physical impersonation. 
\subsection{Attack Scene Object Pose/Translation}
\label{subsec:attack-scene-object-pose}
\begin{wrapfigure}{l}{0.05\textwidth}
    \vspace{-10pt} 
    \centering
    \includegraphics[width=0.08\textwidth]{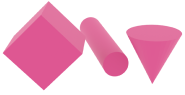}
    \vspace{-25pt} 
\end{wrapfigure}
DNNs are vulnerable to subtle pose or position changes,
with tools like 3DB \cite{leclerc_3db_2021} available for vulnerability exploration. 
Using \diffr, attackers can generate precise object poses or translations to induce misclassification in arbitrary settings. 
Alcorn et al.~[\citeyear{alcorn_strike_2019}] demonstrated that Inceptionv3 misclassifies 97\% of the pose space for ImageNet objects recognized in their canonical poses, with adversarial poses transferring at high rates to AlexNet, ResNet-50, and YOLOv3. 
Similarly, Zeng et al.~[\citeyear{zeng_adversarial_2019}] demonstrated adversarial poses misleading Visual Question Answering models resulting in wrong scene descriptions. 
ViewFool \cite{dong_viewfool_2022} trained NeRF models on 3D objects from BlenderKit\footnote{\url{https://www.blenderkit.com/}}, sampling 100 images per model, and demonstrated that ViT-B/16 was more robust to pose attacks than ResNet-50.
\section{Digital and Physical Attack Domains}
\label{sec:designate-domain}
In this section, we focus on the challenges of creating and evaluating attacks in both digital and physical domains (see Table \ref{tab:attacker-goals-tasks-overview}) and the tools used for attack.
\subsection{Attacks in the Digital Domain}
\label{subsec:attack-digital-domain}
Digital attacks often rely on simulations for controlled testing. 
However, since differentiable renderers (e.g., Mitsuba, PyTorch3D) often lack simulation features, researchers use non-differentiable tools with simulation features instead.
RAUCA \cite{zhou_rauca_2024} and FPA \cite{li_flexible_2024} used Unreal Engine and CARLA, non-differentiable tools that support data capture, diverse scene setups, lighting conditions and self-driving simulations. 
Two other works produced adversarial textures and meshes using PyTorch 3D and then evaluated their robustness within scenes rendered by non-differentiable Blender and Unity tools \cite{zeng_adversarial_2019,meloni_messing_2021}. 
TT3D \cite{huang_towards_2024} created attacked objects using NeRF and then used Blender and Meshlab for testing cross-render transferability.
\subsection{Attacks in the Physical Domain}
\label{subsec:attack-physical-domain}
Implementing real-world adversarial attacks poses several challenges, especially when manufacturing adversarial meshes and textures. 
Post-processing and mesh stabilization may require advanced techniques like Marching Cubes \cite{tu_exploring_2021} to ensure a watertight, non-degenerate mesh that can be 3D printed.
Researchers have also developed flexible 
``universal'' attacks that can be 3D printed once and deployed in multiple scenarios without retraining \cite{abdelfattah_towards_2021}. 
When applying adversarial textures, high-resolution printing or color constraints (Sec. \ref{subsec:attack-scene-texture}) can enhance feasibility; however, covering large surfaces is costly, prompting the use of localized sticker-mode’’ approaches \cite{li_flexible_2024} that only modify a small area (e.g., a vehicle door).
\section{Future Directions}
\label{sec:research-directions}
To further expose DNN vulnerabilities, we propose four directions for \diffr research:
\begin{enumerate}[itemindent=0mm, leftmargin=0pt, label=\textbf{D\arabic*.}, ref=\textbf{D\arabic*}]
\item[] \textbf{Target Diversity.} Many \diffr attacks focus on targeting cars used for autonomous driving.
Meanwhile, use of NeRF and 3D Gaussian Splatting has recently expanded into other real-world applications for robotics and unmanned aerial systems (UAS) but remains largely unconsidered in adversarial ML research.
Exploring more diverse targets in these applications would expand Task \ref{subsec:attack-scene-geometry} and Task \ref{subsec:attack-scene-texture}.
\item[] \textbf{SOTA Models and Other Modalities.} Existing \diffr attacks mainly target image classifiers and object detectors, with limited work on optical flow, depth estimation, point cloud classifiers, and multi-modal or multi-task fusion models \cite{abdelfattah_towards_2021}.
Attacks on 3D scene understanding and advanced tasks like tracking or video recognition remain underexplored, despite the growing use of robust models in robotics and AR. 
Future research could also include newer architectures such as EfficientNet, ViT, and DeiT, which could exhibit different vulnerabilities 
from older models. 
Exploiting these emerging vulnerabilities would advance attacker goals \ref{subsec:attacker-goals-misclassification}–\ref{subsec:attacker-goals-misestimate-depth}.
\item[] \textbf{Attacks Considering Real-World Phenomena.} Current methods use only basic lighting and camera adjustments, such as varying lighting intensity and position or camera resolution. 
This overlooks complex environmental factors (\textit{e.g.}, variable light shapes, shadows, color) and camera parameters (lens-warping, field of view, focus distance, and exposure) that create new attack surfaces in drones and other camera-equipped systems.
Other physical attacks involving placement of lens covers and rolling shutter exploitation \cite{sayles_invisible_2021} are also understudied.
Broadening research on such real-world phenomena would strengthen Tasks \ref{subsec:attack-scene-illumination}, \ref{subsec:attack-scene-sensors}, and \ref{subsec:attack-physical-domain}.
\item[] \textbf{Tools and Pipelines.} While simulators like CARLA are widely used for attack research, \diffr libraries often require specialized knowledge and manual scene configuration. 
Existing GUIs, such as Blender plugins for Mitsuba\footnote{\url{https://github.com/mitsuba-renderer/mitsuba-blender}}, help export scenes but still demand significant expertise in 3D modeling. 
More user-friendly interfaces and integrated pipelines for differentiable renderers would streamline digital attacks (Task \ref{subsec:attack-digital-domain}), ultimately facilitating transfer to physical scenarios (Task \ref{subsec:attack-physical-domain}).
\end{enumerate}
\section{Conclusion}
Understanding the evolving capabilities of \diffr is essential for safeguarding deep neural networks. 
This survey presents a task-guided review of adversarial attacks using \diffr, covering manipulations of 3D objects and scenes that compromise applications like image classification and object detection. 
By categorizing attacker tasks and linking them to goals, we highlight research gaps such as attacks targeting scene parameters (lighting, camera configurations) and the need for user-friendly resources. 
Future work should explore novel attack methods and practical physical evaluations, facilitating more resilient DNN defenses in this rapidly advancing area.
\bibliographystyle{named}
\bibliography{ijcai25}
\end{document}